# A Semantic QA-Based Approach for Text Summarization Evaluation

## Ping Chen, Fei Wu, Tong Wang, Wei Ding

University of Massachusetts Boston
ping.chen@umb.edu

**Abstract**

Many Natural Language Processing and Computational Linguistics applications involve the generation of new texts based on some existing texts, such as summarization, text simplification and machine translation. However, there has been a serious problem haunting these applications for decades, that is, how to automatically and accurately assess quality of these applications. In this paper, we will present some preliminary results on one especially useful and challenging problem in NLP system evaluation – how to pinpoint content differences of two text passages (especially for large passages such as articles and books). Our idea is intuitive and very different from existing approaches. We treat one text passage as a small knowledge base, and ask it a large number of questions to exhaustively identify all content points in it. By comparing the correctly answered questions from two text passages, we will be able to compare their content precisely. The experiment using 2007 DUC summarization corpus clearly shows promising results.

## Introduction

Technologies spawned from Natural Language Processing (NLP) and Computational Linguistics (CL) have fundamentally changed how we process, share, and access information, e.g., search engines, and questions answering systems. However, there has been a serious problem haunting many NLP applications, that is, how to automatically and accurately assess the quality of these applications. In some case, evaluation of a NLP task itself has become an active research area itself, such as text summarization evaluation. The main difficulty for developing such evaluation comes from the diversity of the NLP domain, and our insufficient understanding of natural languages and human intelligence in general. In this paper, we focus on one especially useful and challenging area in NLP evaluation – how to semantically compare the content of two text passages (e.g., paragraphs, articles, or even large corpora). Pinpointing content differences among texts is critical to evaluation of many important NLP applications, such as summarization, text categorization, text simplification, and machine translation. Not surprisingly, many evaluation methods have been proposed, but the quality of existing methods themselves is hard to assess. In many cases, human evaluation must be adopted, which is often slow, subjective, and expensive. In this paper we present an intuitive and innovative idea completely different from existing methods:

*If we treat one text passage as a small knowledge base, can we ask it a large number of questions to exhaustively identify all content points in it?*

By comparing the correctly answered questions from two text passages, we can compare their content precisely. This idea may seem confusing as "circling around the target" instead of "directly hitting the target". However, Our Question Answering (QA)-based content evaluation is intuitive and supported by the following insights:

1) When we assess someone's understanding on a subject, we do not ask him to write down all he knows about the subject. Instead, a list of questions will be asked, and accurate and objective assessment can be achieved by counting the number of correct answers. During this question answering process, we can also identify which areas he needs to improve.

2) Practical operability. When assessing the similarity of two texts, direct comparison may look natural. However, with current methods (no matter supervised or rule-based) this direct approach becomes increasingly difficult as we move to larger text passages. For example, comparing two articles needs to answer the following questions: how to align sentences, how to semantically represent a sentence, how to generate similarity scores without annotated samples (or as few as possible to minimize cost), how to interpret and evaluate these scores, how to find the content differences of two texts, etc.

3) Easy to interpret. Many existing methods only generate a single score, which illustrates little detail as how an assessment measure is generated and offers no help for system improvement. On the other hand, our QA-based approach requires minimum manual efforts, clearly shows how a measure is calculated, and pinpoints exactly the content differences of two text passages.

In next section we will discuss some existing work. Section 3 will show the architecture for our QA-based evaluation approach, and experiment results will be presented in section 4. We will provide some insights and findings when we design our evaluation system and conduct experiments in one discussion section 5. We conclude in Section 6.

## Related Work

Human evaluations of NLP applications are expensive and slow. A fast option is to use crowdsourcing, such as Amazon Mechanical Turk, to quickly get a large amount of evaluation results from non-expert annotators (Callison-Burch 2009, Lasecki 2015). However, besides the expense, there is little control over the annotation quality.

Automatic semantic evaluation has been studied for decades. Evaluation systems like ROUGE (Lin 2004) in summarization and BLEU (Papineni 2002) in Machine Translation, have been widely adopted as de facto measures. Yet these methods utilize only shallow features such as N-grams and longest-common subsequences, which suffer from the inherent term-specificity (Muhalcea 2006). Moreover, they usually require "gold standard" simplifications generated by human annotators as reference, and which are subjective, expensive, and not always available. To better represent semantics, (Mikolov 2013) developed word embedding models (e.g., word2vec) to semantically encode words and phrases. More recent work moves to learn similarity of larger text pieces such as sentences (Le 2014, Kiros 2015). One fundamental difficulty in embedding models is their high requirement of a large number of text samples as targeted text pieces get larger. Secondly, the quality of similarity measures generated by these models is often vague and hard to assess. Usually authors handpick only a few samples, or extrinsic evaluation has to be adopted. For example, (Mueller 2016) shows the following output:

> the similarity of "a boy is waving at some young runners from the ocean" and "a group of men is playing with a ball on the beach" is 3.13 according to the LSTM model and 3.79 according to a dependency tree-based model.

Just by looking at this sentence pair, it is not clear why 3.13 is a better similarity estimation than 3.79. Moreover, current work on text similarity measurement focuses only on generation of such a single similarity score, and largely ignores a more interesting and important issue: **what are the exact content differences between two text passages?**

Although our evaluation method applies to any application where semantic comparison of texts is needed, our current experiments focus on text summarization evaluation. In this section, we will only discuss the existing work on summarization evaluation.

Automatic text summarization is the process to find the most important content from a document and create a summary in natural language. How to automatically evaluate summaries remains a challenging problem (Jones 1995, Jing 1998, Steinberger 2012). Any such process must be able to comprehend the full document; extract the most salient and novel facts; check if all main topics are covered in the summary; and evaluate the quality of the content (Wang 2016). The problem of co-selection measure is that it needs to count the common sentences between a machine summary and one of the human summaries, which introduces a bias since they are based on a small number of assessors, and a small change of sentences may affect the performance. (Donaway 2000) introduced content-based measure: comparing the term frequency (tf) vectors of machine summary with the tf vectors of the full text or human summary. The score is computed based on "bag of words" or "tf-idf" model using cosine similarity. However, it is likely the summary vector is sparse compared with the document vector, and a summary may use terms that are not frequently used in the full document. An alternative is to use Latent Semantic Indexing (LSI) to capture semantic topics based on Singular Value Decomposition (Steinberger 2012). Unfortunately, LSI is expensive to compute, and suffers from the polysemy problem. (Louis 2013) proposed to use input-summary similarity and pseudomodels to assess machine summary without a gold standard. Other content-based measures include Longest Common Subsequence, Unit Overlap (Radev 2002), Pyramid (Nenkova 2004), Basic Elements (Hovy 2006), and Compression Dissimilarity (Wang 2016).

ROUGE (Recall-Oriented Understudy for Gisting Evaluation) is perhaps the most widely adopted automatic summarization evaluation tool. It determines the quality of a summary by comparing it with human summaries using N-grams, word sequences, and word pairs (Lin 2004). Its output correlates very well with human judgements. But ROUGE is unsuitable to evaluate abstractive summarization, or summaries with a significant amount of paraphrasing. (Ng 2015) incorporates word embeddings learned from neural network to ROUGE. There have been other efforts to improve automatic summarization evaluation measures, such as the Automatically Evaluating Summaries of Peers (AESOP) task in TAC. The major problem of these methods is the requirement of "gold standard" summaries, and usually only a single score is generated, which is hard to interpret and does not provide any clue as how a summarization system can be improved.

After examining existing NLP evaluation methods ranging from shallow analysis (e.g., N-gram) to deep semantics (e.g., deep neural network word embeddings), there still remain many major challenges:

- High requirement of manual efforts: in general, supervised machine learning methods (e.g., deep learning or other classification methods) need sufficient annotated samples for robust performance, which become prohibitively expensive when evaluation moves from sentences to large text passages. Even with methods using only shallow features (e.g., ROGUE), gold standard needs to be provided, which is often created by highly trained personnel.

- Lack of details in evaluation: most existing methods produce only a single score as the evaluation result. In evaluation, more information is certainly desirable; and can significantly help researchers gain more insights and improve their work. For example, when simplifying a text passage, it will be very helpful to pinpoint information differences between simplified text and original text, so we will know whether/what information is missing from the simplified version.

In this paper, we will present a question answering-based content evaluation method that can identify information differences of different text passages without any manual efforts. Our method can process various text sizes, ranging from a sentence, a paragraph, a document, to even a large corpus. Due to its fundamental nature, our work can be applied anywhere comparison of two texts is required, including summarization evaluation, text simplification evaluation, and machine translation evaluation.

## A QA-Based method for semantic comparison of texts

Our automated evaluation method will leverage two NLP fields: Question Generation (QG), and Question Answering (QA). Its architecture is illustrated in Figure 1. The main idea is first to generate a large number of questions from an original text passage to exhaustively cover its content, and it is reasonable to assume that the original text contains information to answer these questions. To semantically assess the content of a newly-generated text passage (e.g., a summary, a simplified version, or a translation to another language), a QA system will use the new passage as the only knowledge source to answer the questions generated from the original text. If a question is correctly answered, it means that this new text passage contains the same specific piece of information as in the original text although it may be expressed in a different way. By examining all correct answers, we can have an accurate measure of information contained in the new passage. By comparing the questions that can be answered by original text passage but can not be answered by new passage, we can pinpoint exactly the content differences between these two text passages.

Question generation (QG) has been widely used in many fields. In a document retrieval system, a QG system can be used to construct well-formed questions (Hasan 2012). Current QG methods are designed to generate questions with some focus, e.g. a query in an IR system, main topic in a tutoring system (Le 2014). Usually questions are formed by exploiting named-entity information and predicate-argument structures of sentences. Then a QG system ranks the questions in two aspects. One is the question's relevance to the topic and the subtopics of the original passage; and the other is the syntactic similarity of each question with the original passage. As a result, the system outputs questions with high relevance to the topic of original text passage.

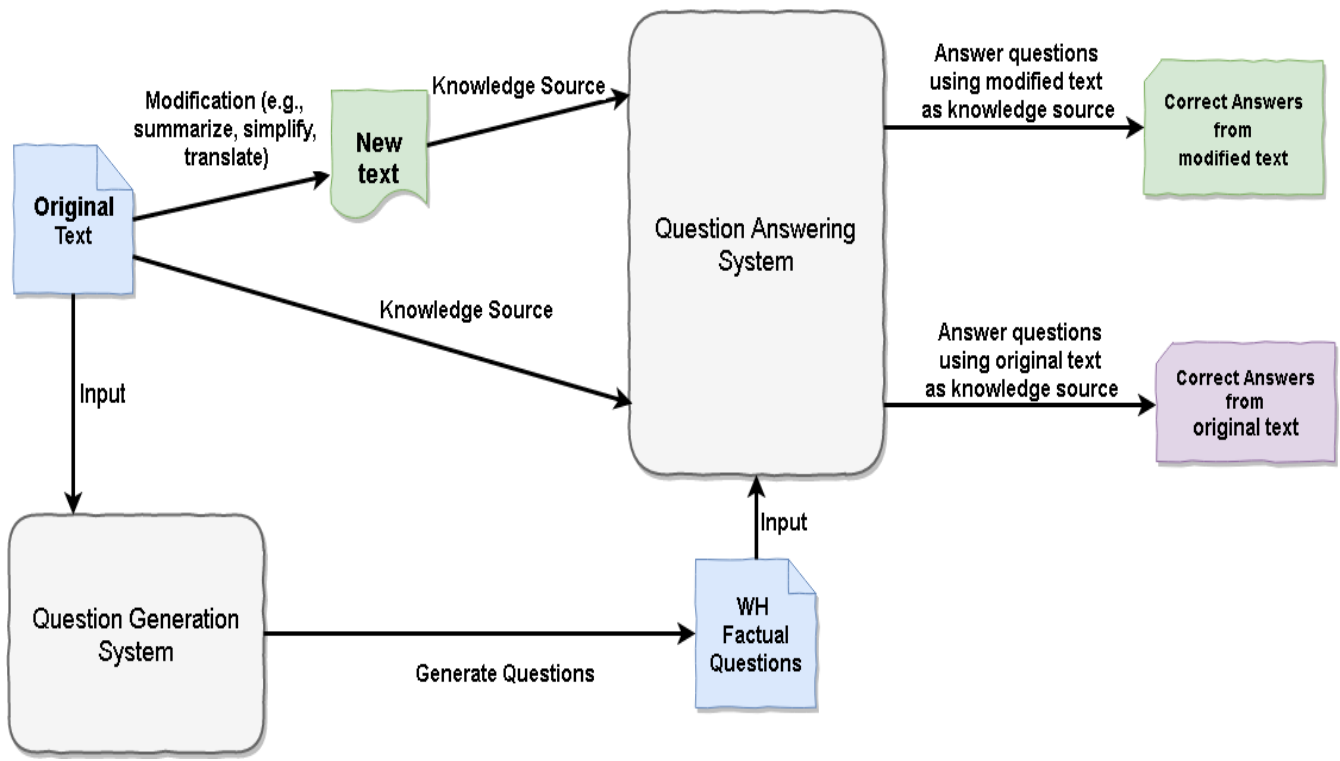

*Figure 1   Our QA-based semantic evaluation system architecture*

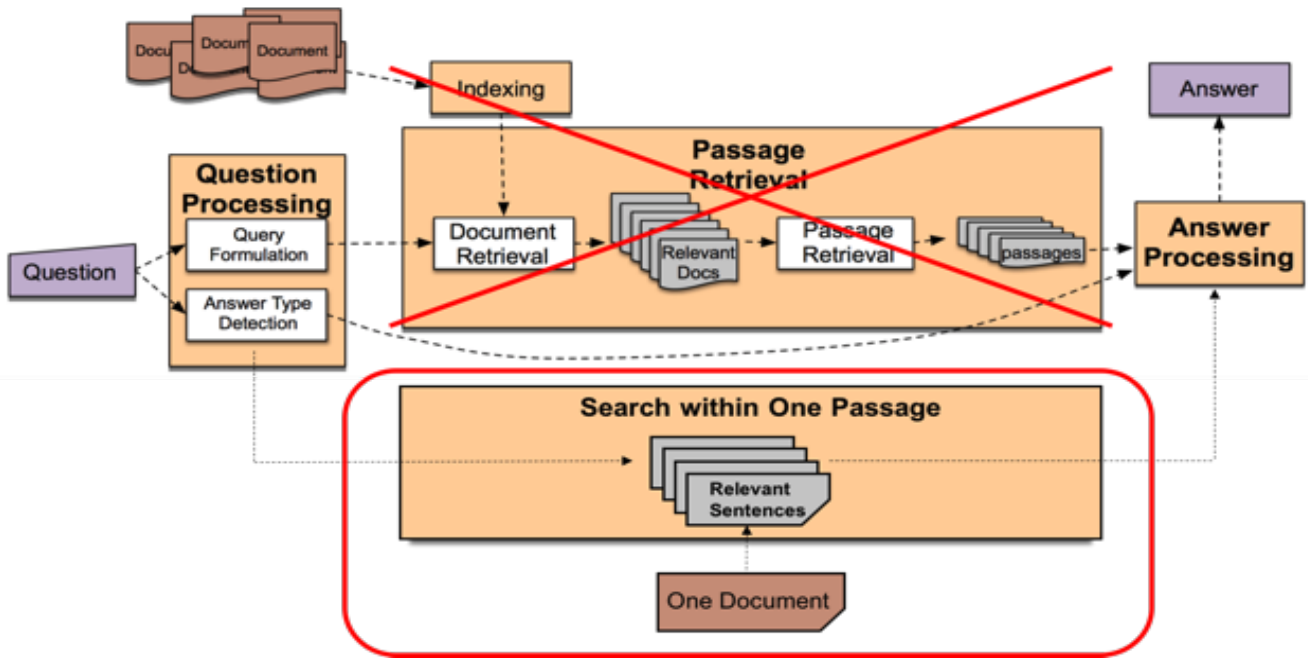

*Figure 2 Customizing a typical QA system for our evaluation approach*

In the QG step we generated factual questions by analyzing the grammatical structure, labeling the lexical items with name entities or other high-level semantic roles (e.g., person, location, time), and performing syntactic transformations such as subject-auxiliary inversion and WH-movement [Heilman 2011]. We used a different ranking component in our QG system so that we could generate not only the questions that closely relate to the topic, but also the questions covering even minor content points. Such questions are highly related to the original text.

On the other hand, these questions might not exclusively cover all the literal information or always be well-structured due to the difficulty in extracting simplified statements from complicated structures in the original text. Hence, using named entities and predefined templates to generate questions can be alternative way in this QG step. We can first apply a Named Entity Recognition method to identify named entities (e.g., person name, time) from a text passage. For each identified named entity, we will generate a set of questions according to predefined question templates. For example, there is one text passage:

*"…… Born in Hodgenville, Kentucky, Lincoln grew up on the western frontier in Kentucky and Indiana……"*

If "Lincoln" is recognized as a person, questions will be automatically generated, e.g.,

- *Who is Lincoln?*
- *When was Lincoln born?*
- *Where was Lincoln born?*
- *When did Lincoln die?*
- *Where did Lincoln die?*

These questions are generated without considering specific text passages, so it is possible some answers cannot be found in the original text passage. In this case, all questions are still asked to an original text passage and to a generated (e.g., summarized, simplified, translated) text passage. The difference of the two answer sets will show the content difference of two texts. The advantage of this approach is that we do not have to always rely on the quality of questions from a QG system. As long as predefined question template is carefully constructed, we can obtain questions with good coverage (over-coverage does not matter) and high quality.

After a large set of questions is generated from original text, we need a QA system to check how many questions can be correctly answered using the content from a single text. A typical QA system usually includes an information-retrieval component to return a large set of ranked documents that may contain the answer. Figure 2 shows the architecture of a customized QA system that will satisfy the needs of this evaluation project. After the question processing step, our QA system component doesn't pass formulated queries to a passage-retrieval component. Instead, it uses the queries to search for relevant sentences within a document, from which the system will extract answers. The change in structure increases difficulties in the question processing step and answer-processing step of the QA system.

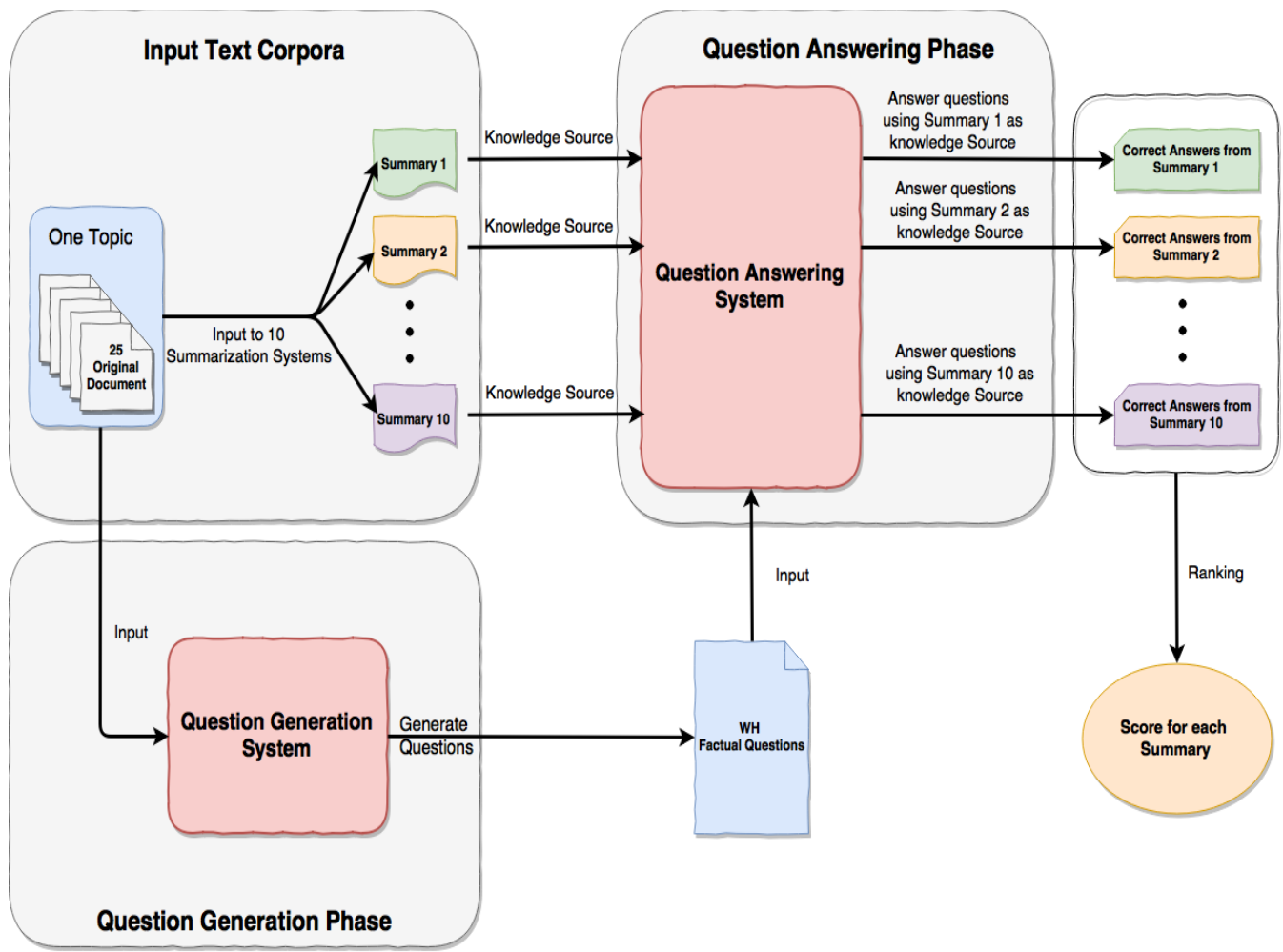

*Figure 3  The QA-based Summarization Evaluation Process Using DUC 2007 corpus*

## Experiment

To test our idea, we have built a proof-of-concept system using some existing QG and QA systems. For the Question Generation component, we adapted Heilman, M. (2011)'s QG system. For the QA component, we need our QA system component to be able to answer questions from a single document, instead of using an information-retrieval system to return a large set of ranked documents that may contain the answer. We take advantage of the open source QA framework, OpenEphyra, by replacing the Passage Retrieval component with a text searching component, which searches within one document.

To test our prototype system, we use the corpus from Document Understanding Conference (DUC) 2007. The corpus contains 2 sets of text passages. The first set is the original documents divided into 45 topics. Each topic consists of 25 original documents. The second set of texts is the summaries of each topic. The summaries were generated by 2 baseline summarization systems, 30 participating summarization systems, and 4 human summarizers. Thus, in total, there are 36 summaries for each topic. All of these summaries have been evaluated by human assessors, and have been given scores on their content responsiveness and linguistic quality.

We hypothesized that the content quality of a summary can be measured by the number of questions answered by the QA system, given this summary as the only knowledge. The whole process of our experiment is shown in Figure 3.

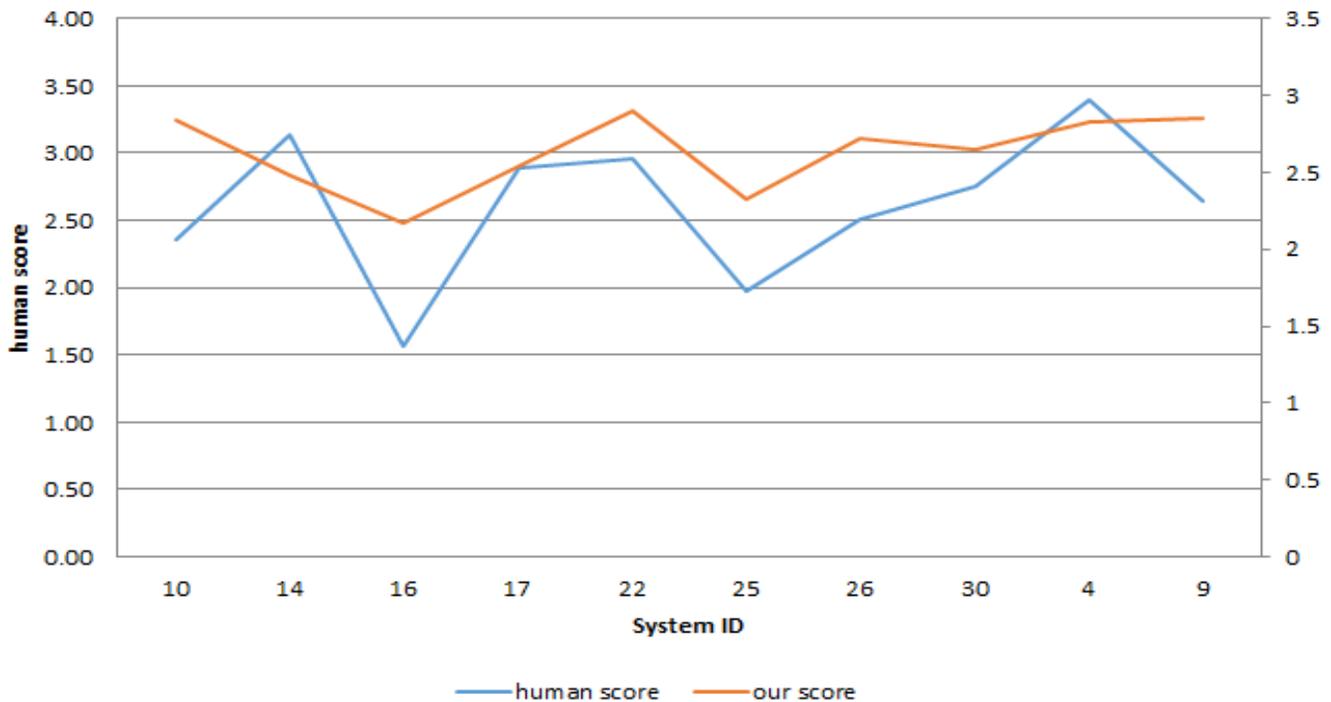

*Figure 4  Comparison of our evaluation scores and human evaluation scores*

In the question generation phase, we use all the 25 original documents of each topic as input to the QG component. As output, the QG component generates a large set of questions for each topic. The number of generated questions ranges from a few hundred to a few thousand, and varies from topic to topic depending on the document length. All questions were limited to "WH" factoid questions that are shorter than a certain threshold. In the QA phase, we run the QA component through each generated summarization from each topic. For summarizations in each topic, we use the set of questions corresponding to this topic as input to the QA system. The goal of our automatic evaluation system is to determine the performance of different automatic summarization systems based on the content quality of the summarizations generated by them. In this experiment, due to time constraints, we chose to compare the performance of 10 summarization systems with system ID 4, 9, 10, 14, 16, 17, 22, 25, 26, and 30. These systems are evenly distributed performance-wise as evaluated by human assessors. To make sure the answers generated by the QA system are mostly correct, we set the confidence score of the answers to a very high value. If the QA system is not highly positive about the answer to a question, it will not answer that question.

As stated above, for each topic, we have a set of questions generated by QG component and these questions are used to evaluate 10 summarization systems' performance over this topic. Each summarization system's overall performance is measured by averaging its performance over the 45 topics.

More specifically, to evaluate 10 summarization systems' performance over a certain topic, we ran the QA system 10 times on this topic. Each time, as input to the QA system we use the same set of questions generated from this topic's original documents, but as knowledge source we use different summaries generated by different summarization systems. Within each run, to measure the performance of a summarization system we calculate the percentage of answered questions, among the total questions for that topic, given this summarization system's summary as the knowledge source. The percentages are later normalized to the range of [1, 5], matching the scores given by the human assessors. Finally, we average the scores of each summarization system over the number of topics. These average scores are the output of our automatic evaluation system, which is the performance measure of each summarization system. In the last step, we compare the content scores given by human assessors, and our system's output score, by computing Pearson's correlation between Automatic Evaluation System scores and human assigned mean content scores. As shown in Figure 4, our performance scores and human score correlate very well.

To further evaluate the robustness of our approach, we varied the parameters used in our system to check how these changes affect the system performance. The confidence threshold was set to a very high value (0.8 or 0.9 as shown in Table 1) to ensure correctness of generated answers and to minimize possibility of generating false answers. As shown in Table 1, in general our scores and human scores correlate very well. The longer questions can generally more difficult for a QA system and sometimes results in lower performance, which is why we limited the question length. Since summarization tends to keep most important information, not all questions are of the same importance. Questions covering important information should be given higher priority. By question filtering, we applied LDA to identify topic words of documents, and filtered out the questions that do not contain these topic words. Performance gets better since a summarization system will not be penalized by not covering unimportant information.

| Confidence | Question Length | Question Filtering | Correlation |
|---|---|---|---|
| 0.8 | 20 | No | 0.6120 |
| 0.9 | 20 | No | 0.6379 |
| 0.8 | 20 | Yes | 0.7725 |
| 0.9 | 20 | Yes | **0.8363** |
| 0.8 | 30 | No | 0.6373 |
| 0.9 | 30 | No | 0.5092 |
| 0.8 | 30 | Yes | 0.6310 |
| 0.9 | 30 | Yes | 0.5462 |
| ROUGE SU4 | | | 0.8827 |
| ROUGE 2 | | | 0.9281 |

**Table 1 Experiment Results**

## Discussion

In this section, we will provide some insights and findings as we design our QA-based semantic evaluation system and analyze the experiment results.

1) We have manually examined some questions that we generated, and found that they are highly related to the text, both semantically and structurally. The over-generating approach helps us to obtain questions that can cover almost all the text content and literal information. The average number of questions generated for a topic with 25 documents is 1193 when setting the question length to be less than 20 words. The large amount of generated questions ensures our system's robustness when a small number of false answers exist.

2) In addition, the generated questions are able to discover basic entity relations within sentences and between sentences. For example, there are questions like "Who were assaulted by Aryan Nations' guards?" "Who does Richard Cohen argue to?" "Who was co-founder of the Southern Poverty Law Center?" "What was Morris Dees the co-founder of?" This fact also suggests the proposed QA based evaluation approach is potentially superior to ROUGE based measures, since ROUGE is not able to find relations between entities. However, our current QG system may fail to generate deep and intelligent inference questions and discover long-distance dependencies.

3) Our method currently focuses more on how much information a summary contains rather than the importance of certain information within the original text by using topic words as a separate filter. Instead if integrating topic words into our QG system, we can provide more personalized evaluation as certain information may be more important to a specific user.

4) The core focus of our idea is on how to semantically compare two texts, which can be a summary and a document, a text and its simplified/revised version, a text and its translation in another language, etc. Hence, our approach can have broad applications in other NLP tasks such as text simplification and machine translation where evaluation is also very important and challenging.

5) Although end-to-end Deep Neural Network methods in NLP become popular due to their good performance, our white-box-style approach has its unique appealing advantages in the evaluation field since humans are often closely involved in this process and need assess the soundness of evaluation and find clues to improve their NLP system.

## Conclusion

In this paper, we present an innovative semantic evaluation method for various NLP applications by leveraging Question Generation and Question Answering fields. Our method requires no manual efforts, is easy to interpret, and illustrates details about NLP systems being evaluated. Our experiments on text summarization evaluation showed promising results. Since our focus is on a fundamental issue in NLP: how to semantically compare two texts, besides summarization evaluation we expect that our idea will have broader applications on various NLP tasks, such as text simplification and machine translation.

## Acknowledgements

This work is partially funded by NSF grant #1241661.